\documentclass[pdflatex,sn-mathphys-num]{sn-jnl}% Math and Physical Sciences Numbered Reference Style 
\usepackage{graphicx}%
\usepackage{multirow}%
\usepackage{amsmath,amssymb,amsfonts}%
\usepackage{amsthm}%
\usepackage{mathrsfs}%
\usepackage[title]{appendix}%
\usepackage{xcolor}%
\usepackage{textcomp}%
\usepackage{manyfoot}%
\usepackage{booktabs}%
\usepackage{algorithm}%
\usepackage{algorithmicx}%
\usepackage{algpseudocode}%
\usepackage{listings}
\usepackage{subfigure}
\usepackage{multirow}
\usepackage{siunitx}
\usepackage{makecell}%
\theoremstyle{thmstyleone}%
\theoremstyle{thmstyletwo}%

\theoremstyle{thmstylethree}%

\begin{document}

\title[Deep-Wide Learning Assistance for Insect Pest Classification]{\centering Deep-Wide Learning Assistance\\ for Insect Pest Classification} 

%%=============================================================%%
%% GivenName	-> \fnm{Joergen W.}
%% Particle	-> \spfx{van der} -> surname prefix
%% FamilyName	-> \sur{Ploeg}
%% Suffix	-> \sfx{IV}
%% \author*[1,2]{\fnm{Joergen W.} \spfx{van der} \sur{Ploeg} 
%%  \sfx{IV}}\email{iauthor@gmail.com}
%%=============================================================%%

\author[1,2]{\fnm{Toan T.} \sur{Nguyen}}\email{nguyentientoan190401@gmail.com}

\author[1,5]{\fnm{Huy T.} \sur{Nguyen}}\email{bihuyz@gmail.com}

\author[4]{\fnm{Huy Q.} \sur{Ung}}\email{xhu-ung@kddi.com}

\author[1]{\fnm{Hieu T.} \sur{Ung}}\email{ungtrunghieu99@gmail.com}

\author*[1,2,3]{\fnm{Binh T.} \sur{Nguyen}}\email{ngtbinh@hcmus.edu.vn}

\affil[1]{\orgname{AISIA Research Lab}, \orgaddress{ \city{Ho Chi Minh City}, \country{Vietnam}}}

\affil[2]{\orgname{University of Science}, \orgaddress{ \city{Ho Chi Minh City}, \country{Vietnam}}}

\affil[3]{\orgname{Vietnam National University}, \orgaddress{ \city{Ho Chi Minh City}, \country{Vietnam}}}

\affil[4]{\orgname{KDDI Research, Inc.}, \orgaddress{\city{Fujimino City}, \country{Japan}}}

\affil[5]{\orgname{Université de Bourgogne}, \orgaddress{ \city{Dijon}, \country{France}}}

%%==================================%%
%% Sample for unstructured abstract %%
%%==================================%%

\abstract{
Accurate insect pest recognition plays a critical role in agriculture. It is a challenging problem due to the intricate characteristics of insects. In this paper, we present DeWi, novel learning assistance for insect pest classification. With a one-stage and alternating training strategy, DeWi simultaneously improves several Convolutional Neural Networks in two perspectives: discrimination (by optimizing a triplet margin loss in a supervised training manner) and generalization (via data augmentation). From that, DeWi can learn discriminative and in-depth features of insect pests (deep) yet still generalize well to a large number of insect categories (wide). Experimental results show that DeWi achieves the highest performances on two insect pest classification benchmarks (76.44\% accuracy on the IP102 dataset and 99.79\% accuracy on the D0 dataset, respectively). In addition, extensive evaluations and ablation studies are conducted to thoroughly investigate our DeWi and demonstrate its superiority. Our source code is available at \href{https://github.com/toannguyen1904/DeWi}{this GitHub repository}.
% \href{https://www.dropbox.com/scl/fo/7qj6zi03u7auh8ubzdtv1/h?rlkey=srezfpubumgpdtrcbqbmtcd1o&dl=0}{ an anonymous repository}.
}

\keywords{Insect Pest Classification, Smart Agriculture, Convolutional Neural Networks, Contrastive Learning, Image Augmentation}

\maketitle

\section{Introduction}\label{Sec:Intro}
Agriculture plays a vital role in the economy of many countries and the food supply for the whole world's population~\cite{johnston1961role,bruinsma2017world}. Many factors, such as climate change, deforestation, soil degradation, and insect pests, affect agricultural crops. For a long time, many studies have been conducted to improve various aspects of agriculture~\cite{liakos2018machine}. Recently, with the great achievements of machine learning, many applications have been developed and practically utilized to improve the farming efficiency and the quality of agricultural produce~\cite{king2017technology, kamilaris2018deep,zhu2018deep, zhang2020applications}. In particular, there are several studies of machine-learning-based methods for weather forecasting~\cite{7415154,wang2012new}, disease forecasting~\cite{domingues2022machine,fenu2019application}, or watering schedule recommendation~\cite{sit2020comprehensive,abioye2022precision}. Following the popularity of high-quality image capturing devices and the rapid development of computer vision algorithms, vision-based solutions have been applied in many aspects of agriculture including monitoring crops~\cite{zhu2016field,sadeghi2017automated}, controlling diseases of crops~\cite{wang2017automatic,sabzi2018fast}, automatic crop harvesting~\cite{wei2018nighttime,davidson2016proof}, and automated management systems for crops~\cite{halstead2018fruit,gonzalez2017development}.

Insect pests are well-known as one of the most significant hazards to agricultural produce~\cite{dent2020insect}. Due to their rapid reproduction, evolution, and spread, insect pests can cause hefty and irreversible damage to crops unless early warnings and actions are taken. However, manually identifying insect pests on a large farm by human experts is extremely time-consuming and expensive. An automated vision-based insect pest recognition system is thus promising to perform this task more efficiently~\cite{rustia2021automatic}. In this work, by training on a large amount of pest image data, our method can accurately classify many pest species given images from different sources (e.g., farmland camera, mobile phone, newspaper, internet) with varying image quality. Furthermore, our model can be potentially embedded in smart devices or camera systems on fields to help farmers identify insect pests effectively. Preventing insect pests in agriculture, however, is a complex and multi-step process. In the scope of this work, we focus on insect pest classification, which is one of the most important and challenging tasks. A high-performance recognition model is of paramount importance for subsequent methods to counter insect pests. These methods include biological control~\cite{kenis2017classical,cock2016trends}, chemical control~\cite{pickett1997developing,roubos2014mitigating}, and insect food consumption~\cite{kuccukkurt2010beneficial,ince2010effect,avci2013effects}.

There are several difficulties in recognizing insect pests. The most notable problem is high intra-class and low inter-class variances. For example, an insect pest species can have different forms at different growing stages (e.g., egg, pupa, worm), while different species could be remarkably similar in visual appearance~\cite{xie2018multi, wu2019ip102}. Therefore, recognizing insect pests can be considered as a fine-grained classification problem that requires specific techniques to handle~\cite{zhao2017survey}. The second challenge is the long-tailed distribution in insect pest data, where some species are far more prevalent than others. Other challenges include the small scales of the insect pests and the missing information caused by other objects (e.g., leaves, soil) obscuring.

Up to now, many methods~\cite{wen2009local, xie2015automatic, wang2012new, ayan2020crop, yang2021convolutional, nanni2022high, yu2022frequency, ung2021efficient, feng2022ms, xia2023ensemble, an2023insect, ali2023faster, guo2024lightweight,ali2023faster,nguyen2024insect}, have been proposed for the insect pest classification task, each focusing on dealing with specific difficulties. For instance, Ung \textit{et al.}~\cite{ung2021efficient} designed an ensemble framework consisting of a feature pyramid network that handles the problem of highly varied scale pests and a multi-branch network that aims at discriminative part localization. For the challenge of long-tailed distribution, Feng \textit{et al.}~\cite{feng2022ms} employed a decoupled learning~\cite{kang2019decoupling} strategy. To address the imbalance data problem, authors in~\cite{zhao2023data} recently proposed a method based on multiple residual convolutional blocks. The limitations of these methods are primarily in the use of ensemble methods that integrate many different models or design complex and burdensome neural networks with many modules. A high-performance and efficient method is the decisive factor for the practicality of insect pest detection systems~\cite{rustia2021automatic}.

With that inspiration, in this research, we propose novel and effective Deep-Wide (DeWi, in short) learning assistance that applies a novel one-stage learning strategy. This strategy enables the model to learn in-depth features by employing the triplet margin loss~\cite{hermans2017defense} and generalize effectively to a wide variety of pest categories with the help of a data augmentation technique - Mixup~\cite{zhang2018mixup}. Additionally, we leverage the discrimination capability of the triplet loss in a novel and effective manner. Our DeWi assistance is simple in design, easy to implement, and compatible with various convolutional Neural Network (CNN) backbone architectures. The main contributions of our work are summarized as follows:

\begin{enumerate}[1.]
\item We design a novel architecture that can apply to multiple CNN backbones. With this architecture, we facilitate the deep-feature learning ability by employing the triplet margin loss in a novel and effective style.

\item We propose an alternating training strategy to simultaneously promote our model's discrimination and generalization capabilities. 

\item We conduct comprehensive experiments and achieve state-of-the-art performances on insect pest classification benchmarks. We also conduct ablation studies to investigate other aspects of our method.
\end{enumerate}

The rest of the paper is organized as follows. First, we briefly present notable related works in insect pest classification, contrastive learning, and data augmentation in Section~\ref{sec: related works}. Next, the details of our proposed DeWi are illustrated in Section~\ref{sec: method}. We then present experiments to demonstrate the effectiveness of our method and compare it with previous techniques in Section~\ref{sec: exps}. Ablation studies are also presented in this section. Finally, our paper ends with our conclusion and discussion for future works in Section~\ref{sec: conclusion}.

\section{Related Works}\label{sec: related works}
This paper draws on the extensive literature of previous works on insect pest classification, contrastive learning, and image data augmentation. In this section, we briefly review the literature and summarize related studies of each domain.

\subsection{Insect Pest Classification}
Insect pest classification has been well-concerned for many years~\cite{li2021classification}. Various traditional methods have been proposed to tackle this task. For instance, Wen \textit{et al.}~\cite{wen2009local} proposed a local feature-based approach for orchard insects employing local region feature detectors for feature detection, the scale-invariant feature transform~\cite{lowe2004distinctive} for local feature description, the bag of words for visual representation, and six classifiers for classification. Xie \textit{et al.}~\cite{xie2015automatic} proposed an insect recognition system using multi-task sparse representation with sparse-coding histogram and multi-kernel learning techniques. The authors in~\cite{wang2012new} constructed an automated identification system by designing seven features following basic geometrical features. The classification results are then obtained using artificial neural networks or support vector machine. Although these traditional methods acquired promising results, they come with several limitations. In particular, they primarily depend on hand-crafted features, which are time-consuming to collect and require professionals' expertise. Traditional approaches, therefore, are easily failed when dealing with domain shifts in the morphological features of insects. Besides, the generalizability of such methods is also not satisfied as the number of insect classes they could handle is severely limited. As a result, performance of traditional methods is considerably low on several insect pest classification benchmarks~\cite{wu2019ip102,kasinathan2021insect}, hindering their applicability in real-world scenarios.

In recent years, Convolutioxnal Neural Networks have been successfully used in multiple domains~\cite{gu2018recent}. CNN-based recognition methods~\cite{simonyan2014very, krizhevsky2017imagenet,howard2017mobilenets,he2016deep,xie2017aggregated,zagoruyko2016wide,li2022image,zhao2023data} have shown great success in different computer vision tasks, e.g., classification~\cite{wang2019development}, detection~\cite{dhillon2020convolutional}, segmentation~\cite{minaee2021image}, and generation~\cite{ho2020denoising}. Modern approaches that leverage CNN models achieve remarkable results in insect pest classification and outperform most traditional techniques. Yang \textit{et al.}~\cite{yang2021convolutional} proposed a convolutional rebalancing network to tackle the problem of long-tailed distribution of insect pest data. Their network includes a convolutional rebalancing module, an image augmentation module, and a feature fusion module. Ung \textit{et al.}~\cite{ung2021efficient} proposed a CNN-based ensemble model including a multi-branch network~\cite{zhang2021multi}, a residual attention network~\cite{wang2017residual}, and a feature pyramid network~\cite{lin2017feature}. Also using the ensemble method, Nanni \textit{et al.}~\cite{nanni2022high} combined six CNNs and optimized the whole framework using two new optimizers improved from the Adam algorithm~\cite{kingma2014adam}. Unlike previous works, our proposed DeWi does not combine many networks or modules. Furthermore, DeWi can be applied or fine-tuned for different CNN backbones. These advantages make DeWi more applicable and adaptable in practical settings.

\subsection{Contrastive Learning}
Contrastive learning has been widely explored in recent years for a variety of fields, such as computer vision~\cite{jaiswal2020survey}, sensor data analysis~\cite{zhou2023self}, natural language processing~\cite{le2020contrastive}, robot learning~\cite{ngyen2023open}, and graph representation learning~\cite{xie2022self}. Contrastive learning consists of strategies that try to learn an embedding space in which the embeddings of similar samples stay close to each other while the embeddings of dissimilar ones are far apart. A similarity metric is usually employed to determine how close two embeddings are. A contrastive loss is the critical component of most contrastive learning methods. In the particular field of computer vision tasks, it is computed on the embeddings of the images extracted from an encoder network. The model then learns meaningful representations of the input images by optimizing that loss. The obtained model weights are later transferred to multiple downstream tasks.

Contrastive learning usually utilizes contrastive pairs or triplets to achieve generalizable and transferable representations. For example, the authors in~\cite{oord2018representation} proposed a framework that encodes predictions over future observations by extracting compact latent representations and optimizing a loss function based on noisy-contrastive estimation~\cite{gutmann2010noise} called InfoNCE. A dynamic memory bank was presented by He \textit{et al.}~\cite{he2020momentum} to store the embeddings of negative sample data. To make it more challenging to distinguish negative samples from positive queries, Hu \textit{et al.}~\cite{hu2021adco} used adversarial training that maximizes the adversarial contrastive loss of miss-assigning each positive sample to negative ones. Other studies~\cite{caron2020unsupervised,li2020prototypical} suggested using prototypical representations to discriminate the data to locate more appropriate negative samples. Chen \textit{et al.}~\cite{chen2020simple} demonstrated the importance of large batch size and the number of training steps in contrastive self-supervised learning. Zbontar \textit{et al.}~\cite{zbontar2021barlow} proposed a loss function that naturally avoids trivial solutions~\cite{jing2022understanding} by calculating the cross-correlation matrix between the outputs of two identical networks fed with distorted versions of a sample, and making it close to the identity matrix. Inspired by the success of contrastive learning in many problems, in this work, we apply a contrastive learning loss function - the triplet margin loss~\cite{hermans2017defense} in a novel one-stage manner. Experimental results indicate that this approach significantly improves our model's ability in learning meaningful representations while optimizing for the primary classification task.

\subsection{Image Augmentation}
Insect pests can take on various appearances. Applying image augmentation methods is therefore a promising approach to improve the generalizability of recognition models. Two traditional directions of image augmentation are geometric transformation and color processing~\cite{shorten2019survey}. With geometric transformation, an image can be rotated to change the perspective of its objects. Besides, depending on the characteristics of the training and testing sets, flipping can be performed on images horizontally or vertically.
Color image processing for augmentation assumes that the distributions of the training and testing datasets differ in colors, such as contrast or lighting. Such methods alter the color properties of the input image by changing its pixel intensity.
These traditional augmentation methods preserve the input image's general structure and the corresponding label, thus limiting the model's generalization ability~\cite{xu2023comprehensive}.

Over the years, many advanced image augmentation methods have been proposed to improve the performance of deep neural networks on imagery data. Intensity transformation is a noteworthy sub-domain. One of the soonest and most straightforward strategies in this kind is random noise injection, such as Gaussian noise~\cite{vapnik1991principles}. One sub-area of intensity-based data augmentation is image mixing and image deleting. To encourage the network to focus on the overall structure of the object, this particular sort of augmentation either fuses two images or deletes parts of an image to obscure or confuse specific image properties. CutOut~\cite{devries2017improved}, CutMix~\cite{yun2019cutmix}, and MixUp~\cite{zhang2018mixup} are notable techniques that achieved significant improvements with various CNN architectures. With these advanced methods, objects of different classes can be mixed in a single image, and the one-hot label is no longer preserved. Among those methods, Mixup has shown to be the most impactful in the contrastive learning field. There have been many recent additional studies of Mixup, such as Manifold MixUp~\cite {verma2019manifold} or MixCo~\cite{kim2020mixco}.
In this study, we leverage this method for our DeWi assistance to enhance the generalization abilities of CNN models.

\section{Methodology}\label{sec: method}

\begin{figure}[h]
	\centering
	\includegraphics[width=1\linewidth]{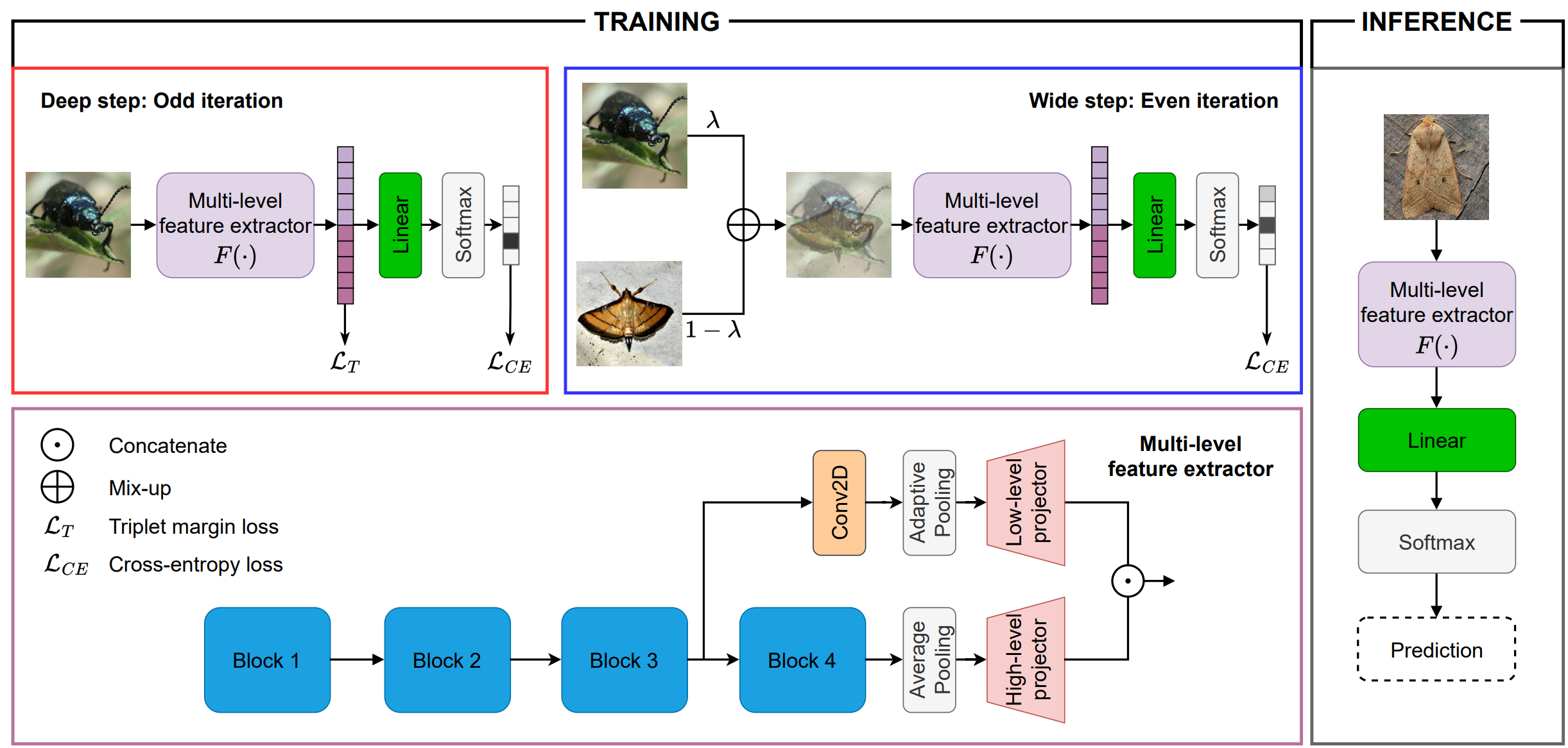}
	\caption{The overview of our proposed method. During training, the Deep step and Wide step are alternately applied in an epoch. We use the same network architecture in Deep and Wide steps. The triplet margin loss is computed in the Deep step, while the Mixup data augmentation is employed in the Wide step. In inference, the input image is fed through the entire network, from the multi-level feature extractor $F(\cdot)$ to the linear layer and the softmax layer, to get the final prediction.}
	\label{fig: overview}
\end{figure}

In this section, we present our novel DeWi learning assistance to assist residual-network-based classification models (ResNet-based models, in short) during their training process. To improve discrimination and generalization capabilities, DeWi consists of Deep steps and Wide steps, which are alternately applied during training, as shown in Figure~\ref{fig: overview}. The architecture of our model consists of three main modules: a multi-level feature extractor, two loss functions, (i.e., the traditional cross-entropy loss~\cite{good1952rational} and the triplet margin loss~\cite{hermans2017defense}), and the Mixup data augmentation module~\cite{zhang2018mixup}. Our proposed model is trained by the supervised learning strategy iteration-by-iteration \cite{murphy2012machine}. In an epoch, an iteration is defined as a process of forward propagation followed by backward propagation for the mini-batch of images. The Deep step is executed at the $(2e+1)$-th ($e\in \mathbb{N}$) iterations (odd iterations), whereas the Wide step is executed at the $(2e)$-th iterations (even iterations). Two steps cannot be applied in the same iteration because the triplet loss in the Deep step requires labels of training samples as one-hot vectors, while those in the Wide step are mixed labels due to applying the Mixup data augmentation.

We design DeWi for the ResNet-based models (ResNet~\cite{he2016deep}, ResNext~\cite{xie2017aggregated} and Wide ResNet~\cite{zagoruyko2016wide}), but the method can be adapted for use with other networks. We describe our architecture and learning strategy in detail below.

\subsection{Deep step}\label{subsec: project net and triplet loss}
We first define the important notations. Given a mini-batch of $B$ input samples, we denote the $t$-th training sample as $\left (\mathbf{x}^{(t)}, \mathbf{y}^{(t)}  \right )$, $t = 1,2,..,B$, where $\mathbf{x}^{(t)}\in \mathbb{R}^{H\times W\times C}$ is the $C$-channel input image of size $H\times W$ and $\mathbf{y}^{(t)} \in \mathbb{R}^{K}$ is the one-hot ground truth label, $K$ is the number of insect pest classes in the dataset. We also denote $\hat{\mathbf{y}}^{(t)} \in \mathbb{R}^{K}$ as the results from the classifier. Inspired by contrastive learning, we propose the Deep training step to assist the ResNet-based models in extracting fine-grained and meaningful features. In the Deep step, we apply an additional contrastive loss function, i.e. the triplet margin loss, to force these models to discriminate images that are similar in appearance but belong to different classes.

 In order to apply the triplet margin loss~\cite{hermans2017defense}, we modify the ResNet-based networks via the following steps. First, we remove the final linear layer from these networks. We then form a multi-level feature extractor $F(\cdot)$ by attaching two projectors to these ResNet-based models. Concretely, a high-level projector is connected to an average pooling layer right after the last core block (it is important to know that the ResNet-based models have a general architecture of four consecutive core blocks - blue blocks in Figure~\ref{fig: overview}). Besides, a low-level projector is attached to the end of the third core block to encourage the model to learn fine-grained features at a lower level, as shown in Figure~\ref{fig: overview}. To connect the low-level projector to the third core block, the output of this block is first plugged into a 2D convolution layer of 2048-dimensional (dim) output. Its output is then fed to an adaptive pooling layer, whose output is flattened as a 2048-dim feature vector. This vector is then used as input to the low-level projector.

\begin{figure}[h]
	\centering
	\includegraphics[width=0.85\linewidth]{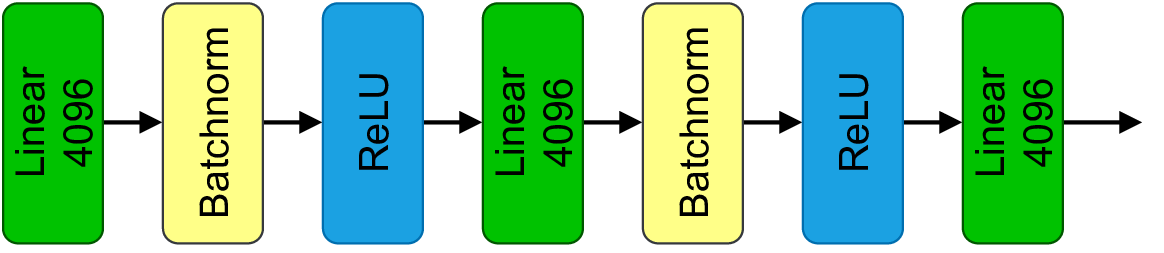}
	\caption{The detailed architecture of our projector network. Both the high-level and low-level projectors produce 4096-dim vectors.}
	\label{fig: projector}
\end{figure}

Both the low-level and high-level projectors have the same architecture, as shown in Figure~\ref{fig: projector}; however, their parameters are not shared. Each projector has three linear layers of 4096-dim output. The first two layers are followed by a batch normalization layer~\cite{ioffe2015batch} and ReLU activation function~\cite{fukushima1975cognitron}. For each input image $\mathbf{x}^{(t)}$, we obtain two 4096-dim embedding vectors from each projector. We then concatenate these two vectors to get a multi-level semantic representation vector $\mathbf{z}^{(t)}$ of 8192-dim.

Our main objective in this design is to optimize an embedding space where the learned feature vectors of images of the same class are close together, and features of images of different classes are located far apart. To achieve that, we optimize our model using the triplet margin loss~\cite{hermans2017defense} as follows:
\begin{equation}
\label{eq: triple loss}
\resizebox{0.93\textwidth}{!}{
$
  \mathcal{L}_{T}(\{\mathbf{z}^{(t)}, \mathbf{y}^{(t)}\}_{t=1}^B) = \sum_{t=1}^{B}\left [ m + \max_{\begin{matrix}
p=1,...B\\ 
\mathbf{y}^{(p)}=\mathbf{y}^{(t)}
\end{matrix}}D\left ( \mathbf{z}^{(t)}, \mathbf{z}^{(p)} \right ) - \\ \min_{\begin{matrix}
n=1,...B\\ 
\mathbf{y}^{(n)}\neq \mathbf{y}^{(t)}
\end{matrix}}D\left ( \mathbf{z}^{(t)}, \mathbf{z}^{(n)} \right ) \right ]_{+},
$
}
\end{equation}
where the margin $m$ is a hyper-parameter, and $D(\cdot)$ is a distance function; we use the Euclidean distance in this study. As presented in Equation~\ref{eq: triple loss}, we apply the batch hard mining strategy introduced in~\cite{hermans2017defense} for our triplet margin loss. Given the semantic representation of one sample, we find the hardest positive sample (of the same class but with the largest distance) and the hardest negative sample (of a different class but with the smallest distance) and use those to form a single triplet contributing to the loss function.

Subsequently, we normalize the embedding $\mathbf{z}^{(t)}$ by a drop-out layer and feed it to a linear layer, followed by a softmax layer, to get the final outputs $\hat{\mathbf{y}}^{(t)}$. During the Deep step, we optimize the overall loss function formulated as follows:
\begin{equation}
\resizebox{0.93\textwidth}{!}{
$
  \mathcal{L}(\{\mathbf{z}^{(t)}, \hat{\mathbf{y}}^{(t)}, \mathbf{y}^{(t)}\}_{t=1}^B) = \beta_1 \times \mathcal{L}_{CE}(\{\hat{\mathbf{y}}^{(t)}, \mathbf{y}^{(t)}\}_{t=1}^B) + \beta_2 \times \mathcal{L}_{T}(\{\mathbf{z}^{(t)}, \mathbf{y}^{(t)}\}_{t=1}^B),
$
}
\end{equation}
where $\mathcal{L}_{CE}$ is the cross-entropy loss~\cite{good1952rational} computed on the final outputs and the ground-truth labels, and $\beta_1$ and $\beta_2$ are hyper-parameters.

\subsection{Wide step}\label{subsec: mixup}
\begin{figure}[h]
	\centering
	\includegraphics[width=0.7\linewidth]{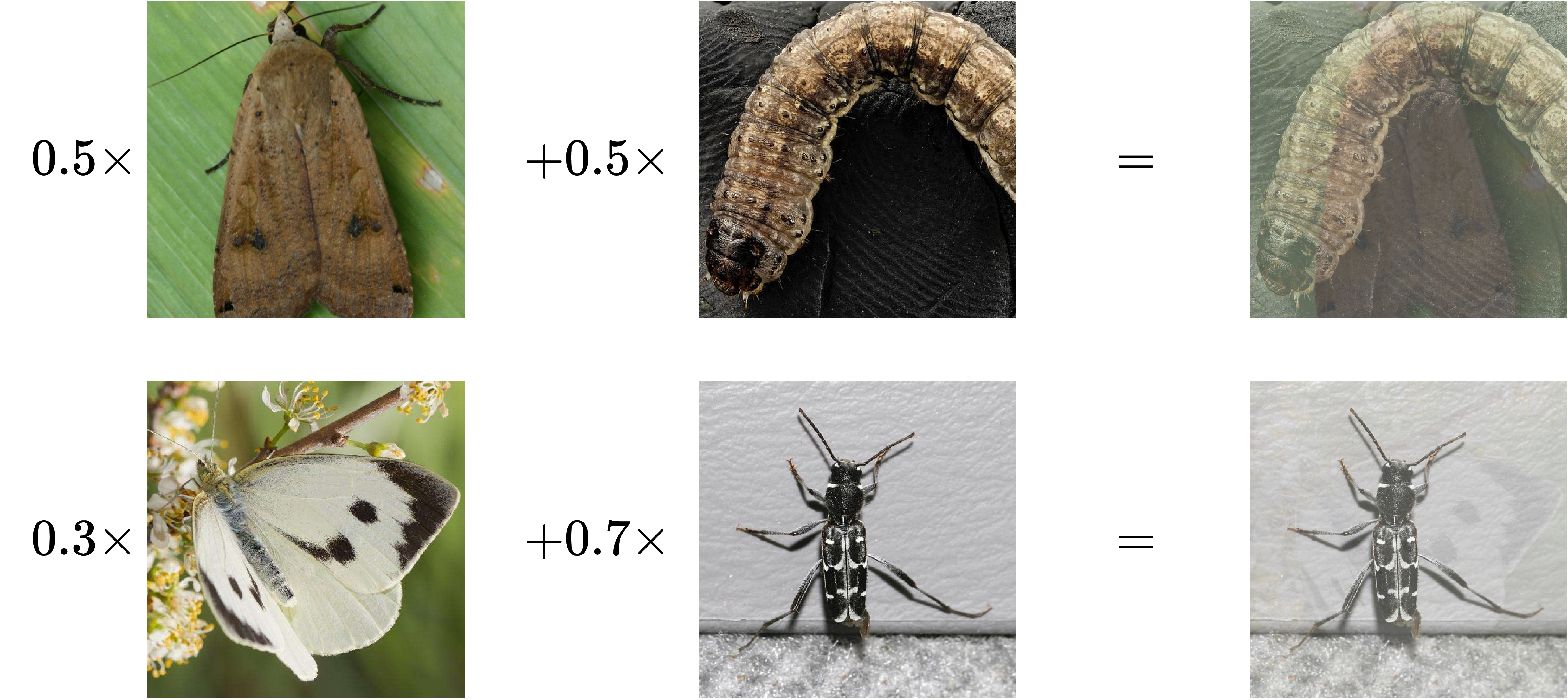}
	\caption{Examples of images from IP102 dataset augmented by Mixup. \textit{Top:} mix of two images of different classes. \textit{Bottom:} mix of two images of the same class.}
	\label{fig: mixup}
\end{figure}

In the Wide step, our assistance applies the Mixup data augmentation~\cite{zhang2018mixup} for images in a batch. A new input image and the corresponding label are sampled from two randomly chosen original samples $\left (\mathbf{x}^{(i)}, \mathbf{y}^{(i)}  \right )$ and $\left (\mathbf{x}^{(j)}, \mathbf{y}^{(j)}  \right )$ as follows:
\begin{align}
\tilde{\mathbf{x}} &= \lambda\mathbf{x}^{(i)} + \left ( 1-\lambda \right )\mathbf{x}^{(j)}, \\
\tilde{\mathbf{y}} &= \lambda\mathbf{y}^{(i)} + \left ( 1-\lambda \right )\mathbf{y}^{(j)},
\end{align}
where $\lambda\sim \text{Beta}(\alpha, \alpha)$, for $\alpha\in\left (0, \infty \right )$. Examples of Mixup augmentation are shown in Figure~\ref{fig: mixup}. It is important to note that we do not increase the number of images loaded to memory. The new images are sampled from the original ones, but the overall number of samples fed to the network stays the same. From that, we do not burden the learning process, yet we still utilize the advantages of Mixup augmentation. We use the same architecture in the Wide step as in the Deep step. Yet in this step, we optimize the cross-entropy loss $\mathcal{L}_{CE}$ alone with new ground truth $\tilde{\mathbf{y}}$ obtained since the triplet margin loss requires the initial one-hot labels, which are not preserved when applying Mixup augmentation. The pseudo-code for DeWi in a particular epoch is shown as Algorithm~\ref{alg: dewi}.

In inference, the input image is fed
through the whole network (the multi-level feature extractor, then the linear layer, and
finally, the softmax layer) to predict the insect pest class. The inference stage of DeWi is shown in Figure~\ref{fig: overview}.

\begin{algorithm}[h]
   \caption{Pseudo-code for DeWi in the training phase.}
   \label{alg: dewi}
    \definecolor{codeblue}{rgb}{0.25,0.25,0.75}
    \lstset{
      basicstyle=\fontsize{7.5pt}{7.5pt}\ttfamily\bfseries,
      commentstyle=\fontsize{7.5pt}{7.5pt}\color{codeblue},
      keywordstyle=\fontsize{7.5pt}{7.5pt},
    }
\begin{lstlisting}[language=python]
# INPUT: 
#  + batch_data: a batch of input images.
#  + batch_labels: a batch of corresponding labels.
# OUTPUT: classification model trained.

# NOTATIONS:
# + mnb_data: a mini-batch of input images in batch_data.
# + mnb_labels: a mini-batch of corresponding labels in batch_labels.
# + F: the multi-level feature extractor.
# + Linear: linear layer.
# + Softmax: softmax function.
# + l_triplet: triplet loss function with the batch hard miner.
# + l_ce: cross-entropy loss function.
# + mixup: mixup function.
# + beta1, beta2: weights of cross-entropy and triplet loss values.

Initialize deep_turn to True

Repeat for (batch_data, batch_labels):
    If deep_turn is True:
        # Deep step
        For each (mnb_data, mnb_labels) in (batch_data, batch_labels):
            features = F(mnb_data)
            features = Linear(features)
            results = Softmax(features)
            triplet_loss = l_triplet(features, mnb_labels)
            ce_loss = l_ce(results, mnb_labels)
            total_loss = beta1 * ce_loss + beta2 * triplet_loss

            Update model using total_loss
    Else:
        # Wide step
        For each (mnb_data and mnb_labels) in (batch_data, batch_labels):
            mixed_data, mixed_labels = mixup(mnb_data, mnb_labels)
            features = F(mixed_data)
            features = Linear(features) 
            results = Softmax(features)
            ce_loss = l_ce(results, mixed_labels)
            total_loss = ce_loss

            Update model using total_loss

    If stopping training conditions are satisfied:
        Stop repeat.

    Toggle deep_turn

End Repeat
\end{lstlisting}
\end{algorithm}

\section{Experiments}\label{sec: exps}
In this section, we conduct several experiments to validate the effectiveness of DeWi. First, DeWi is compared with other state-of-the-art methods on the two widely used insect pest classification benchmarks, i.e., IP102~\cite{wu2019ip102} and D0~\cite{xie2018multi}. Afterward, we analyze the performances of baseline residual models with/without applying our DeWi.

\subsection{Datasets}
\begin{figure}[h]
	\centering
	\includegraphics[width=0.7\linewidth]{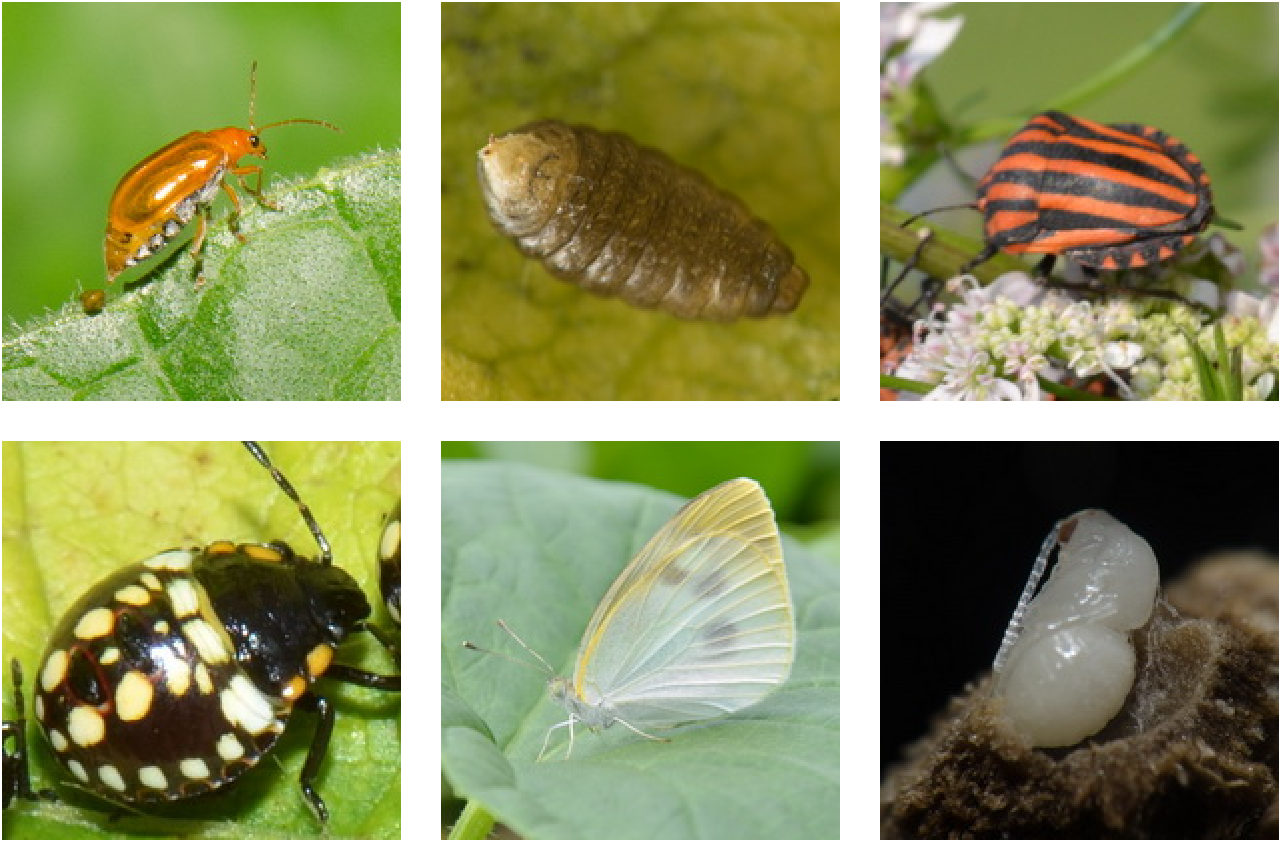}
	\caption{Examples of six insect pests in D0 dataset.}
	\label{fig: d0}
\end{figure}

\begin{figure}[h]
	\centering
	\includegraphics[width=0.9\linewidth]{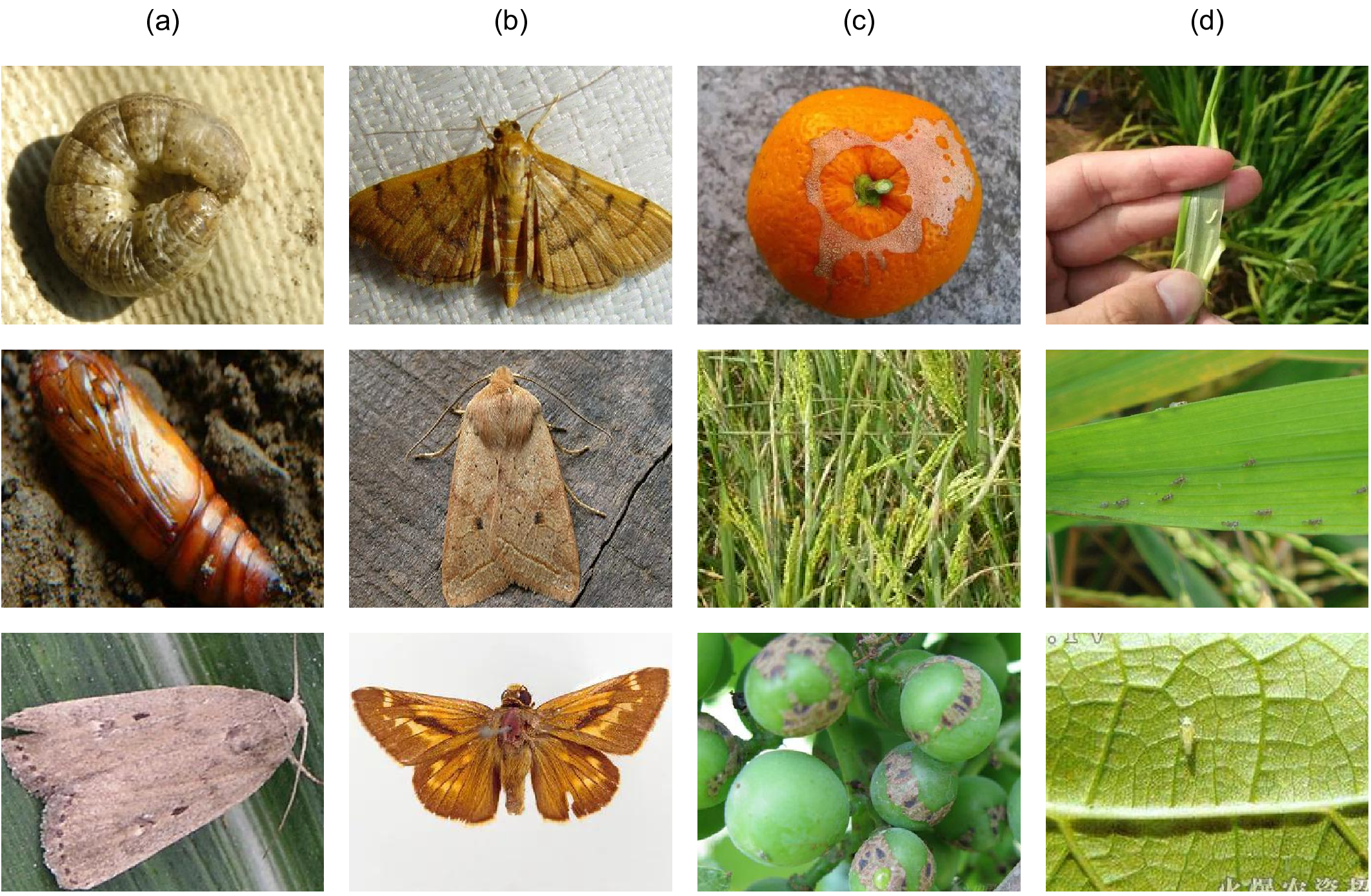}
	\caption{Samples of IP102 dataset. Column \textit{(a)} presents three different morphologies corresponding to three developmental stages of the same worm species. Column \textit{(b)} shows examples of three different butterfly species, but their appearances are particularly hard to distinguish. Column \textit{(c)} gives examples of damaged crop fields without the appearance of insect pests. Column \textit{(d)} shows images of small-scale insects on noisy backgrounds.}
	\label{fig: ip102}
\end{figure}

There are several datasets that have been introduced for the task of insect pest recognition~\cite{samanta2012tea,venugoban2014image,wang2012new,xie2018multi,xie2015automatic,wu2019ip102}. Among them, the D0~\cite{xie2018multi} and IP102~\cite{wu2019ip102} datasets are ones with the largest numbers of insect classes. Therefore, we use them to evaluate our proposed method in this study. D0 contains 4,508 images of 40 insect pest species captured in the natural environment. Figure~\ref{fig: d0} shows examples of this dataset.
IP102 is currently the largest public dataset for the insect pest classification task, with 75,222 images of 102 pest species. Examples of IP102 are shown in Figure~\ref{fig: ip102}. This dataset is more challenging than D0 due to multiple factors. Firstly, the challenge of high intra-class (shown in column \textit{(a)} of Figure~\ref{fig: ip102}) and low inter-class (shown in column \textit{(b)} of Figure~\ref{fig: ip102}) variances of its is particularly significant. Secondly, there are a considerable number of images captured of the damaged crops caused by pests, as shown in column \textit{(c)}. In addition, there are images of small-scale pests on noisy backgrounds, as shown in column \textit{(d)}. Lastly, the distribution of instances of different classes in IP102 is significantly imbalanced. This dataset also covers a large number of pest species and a massive number of images, providing a practical simulation of the biodiversity of pests in real-world scenarios.

\subsection{Evaluation metrics}
Previous works~\cite{ung2021efficient,feng2022ms} primarily evaluated the insect recognition models on three metrics, consisting of the accuracy score (Acc), the macro-average F1-score (mF1), and the geometric mean score (GM). Similarly, we also use these metrics in our experiments. We detail their calculations in detail below.

We compute the recall (Rec) for each class and then take the average over all classes to obtain the mean recall (mRec) as follows:
\begin{equation}
  \text{Rec}_k=\frac{\text{TP}_k}{\text{TP}_k+\text{FN}_k},
\end{equation}
\begin{equation}
  \text{mRec}=\frac{\sum_{k=1}^K\text{Rec}_k}{K},
\end{equation}
where $k$ denotes the $k$-th class and $K$ is the number of classes. TP and FN stand for the true positive and the false negative outcomes, respectively. Similarly, the precision (Pre) for one class and the mean precision (mPre) is defined by:
\begin{equation}
  \text{Pre}_k=\frac{\text{TP}_k}{\text{TP}_k+\text{FP}_k},
\end{equation}
\begin{equation}
  \text{mPre}=\frac{\sum_{k=1}^K\text{Pre}_k}{K}.
\end{equation}
FP stands for the false positive result. mF1 is then computed as the harmonic mean of mRec and mPre as follows:
\begin{equation}
\text{mF1}=2\times\frac{\text{mRec}\times\text{mPre}}{\text{mRec}+\text{mPre}}.
\end{equation}

The accuracy is computed by the number of true predictions over all samples:
\begin{equation}
\text{Acc}=\frac{\sum_{k=1}^K\text{TP}_k}{N},
\end{equation}
where $N$ is the number of image samples.

Finally, we calculate the geometric mean score from the recall outcomes of every class as follows:
\begin{equation}
\text{GM}=\prod_{k=1}^K\sqrt[K]{\text{Rec}_k}.
\end{equation}

\subsection{Experimental settings}
The first dataset used in this work is the IP102 dataset. The authors in~\cite{wu2019ip102} partitioned the IP102 into three subsets: a training set of 45,095 images, a validation set of 7,508 images, and a testing set of 22,619 images. For fair comparisons, we used the same setting of IP102 in our experiments.
All IP102 images are first resized to $400 \times 400$. In the training phase, the images are applied multiple transformations after being randomly cropped to $384 \times 384$. For validation and testing, the images are center-cropped to $384 \times 384$.
For the D0 dataset, we apply the ratio 7:1:2 for training, validation, and testing sets, which is the same as in ~\cite{ung2021efficient}. We do not crop the validation and test images of this dataset.

We initialize residual backbones using the weights pre-trained on the Imagenet dataset~\cite{deng2009imagenet}. We set $\alpha=1$ for Mixup augmentation. For the hyper-parameter $m$ of the triplet margin loss, we try multiple values of 0.1, 0.2, and 0.3, among which 0.2 gives the best performance. Weights of the loss function in the Deep step are set as $\beta_1=\beta_2=1.0$. During training, we optimize the model using Stochastic Gradient Descent (SGD)~\cite{ruder2016overview} with a momentum of 0.9 and an L2 weight decay of $1\mathrm{e}{-4}$. We use multi-step decay for scheduling the learning rate after every 15 epochs with an initial learning rate of $3\mathrm{e}{-3}$, a decay rate of 0.9, and the minimum learning rate of $3\mathrm{e}{-7}$. The mini-batch size and the number of epochs are set to 32 and 100, respectively.

\subsection{Compare with other state-of-the-art methods}\label{sec:sota compare}
We compare our proposed method with other state-of-the-art insect pest classification methods on IP102 and D0 datasets. For brevity, we refer to the networks with the support of DeWi as DeWi networks or DeWi models, or just DeWi. We use ResNet-152 for the backbone of DeWi. The comparison results are shown in Table~\ref{tab: ip102 sota compare}. On IP102, DeWi outperforms other methods by large margins on all three metrics. Specifically, DeWi outperforms the second-best method by 1.83, 1.64, and 1.75 percentage points on accuracy, mF1, and GM, respectively. On the D0 dataset, DeWi also achieves the highest results. Concretely, DeWi reaches 99.79\% accuracy while the second place method~\cite{ung2021efficient} obtains 99.78\%.

\begin{table}[h]
\caption{Comparison between our proposed DeWi and the previous works.}
\label{tab: ip102 sota compare}
\vskip 0.15in
\begin{tabular}{llccc}
\toprule
Dataset & Method  &  Acc & mF1 & GM \\
\midrule
IP102 & Ayan \textit{et al.}~\cite{ren2019feature} & 67.13 & 65.76 & - \\
& Yang \textit{et al.}~\cite{yang2021convolutional} & 70.42 & - & - \\
& Nanni \textit{et al.}~\cite{nanni2022high} & 73.62 & - & - \\
& Ung \textit{et al.}~\cite{ung2021efficient} & 74.13 & 67.65 & 62.52 \\
& Feng \textit{et al.}~\cite{feng2022ms} & \underline{74.61} & \underline{67.82} & \underline{63.32} \\
& DeWi (ours) & \bf{76.44} & \bf{69.46} & \bf{65.07} \\
\midrule
D0 & Thenmozhi \textit{et al.}~\cite{thenmozhi2019crop} & 95.97 & - & - \\
& Yu \textit{et al.}~\cite{yu2022frequency} & 98.16 & 98.34 & - \\
& Ayan \textit{et al.}~\cite{ayan2020crop} & 98.81 & 98.88 & - \\
& Ung \textit{et al.}~\cite{ung2021efficient} & \underline{99.78} & \underline{99.68} & \underline{99.65} \\
& DeWi (ours) & \bf{99.79} & \bf{99.70} & \bf{99.65} \\
\end{tabular}
\vskip -0.1in
\end{table}

\subsection{Compare with baseline residual networks}
\label{sec:baseline compare}
We investigate the improvements of DeWi-assisted models compared to baseline residual networks. We also validate the effectiveness of the multi-level extractor by comparing DeWi models with their counterparts that use a single-level projector.
In the design of such models, we only use the high-level projector and discard the low-level one. We change the output dimension of the last linear layer of the high-level projector to 8192 and compute the triplet margin loss for its output vector. The rest of the network stays the same.

\begin{figure}[h]
	\centering
	\includegraphics[width=0.9\linewidth]{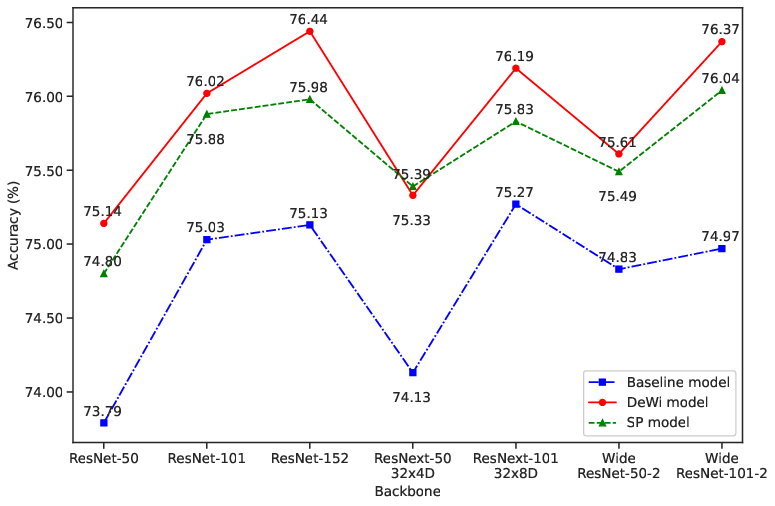}
	\caption{Accuracy score comparison between original residual networks, DeWi networks, and networks using a single projector (\emph{SP models}). We use different residual models for our evaluation, i.e., ResNet, ResNext, and Wide-ResNet.}
	\label{fig: baseline compare}
\end{figure}

We report the results on the accuracy score in Figure~\ref{fig: baseline compare}. From now on, the IP102 dataset will be used for our evaluation if not stated otherwise. The results show that DeWi significantly improves the performances of all the baseline networks, from 0.78 percentage points on Wide ResNet-50-2 to 1.4 percentage points on Wide ResNet-101-2. Additionally, the reported results demonstrate the effectiveness of the design of the multi-level feature extractor when models applying this design (DeWi models) outperform their single-projector counterparts. On our best version using ResNet-152 as the underlying backbone, DeWi outperforms by 0.46 percentage points in accuracy. These results demonstrate that the multi-level extractor helps models learn important features at both low and high levels. In addition, the two projectors support each other and contribute to the overall performance.

\subsubsection{Activation map}
\begin{figure}[h]
	\centering
	\includegraphics[width=1.0\linewidth]{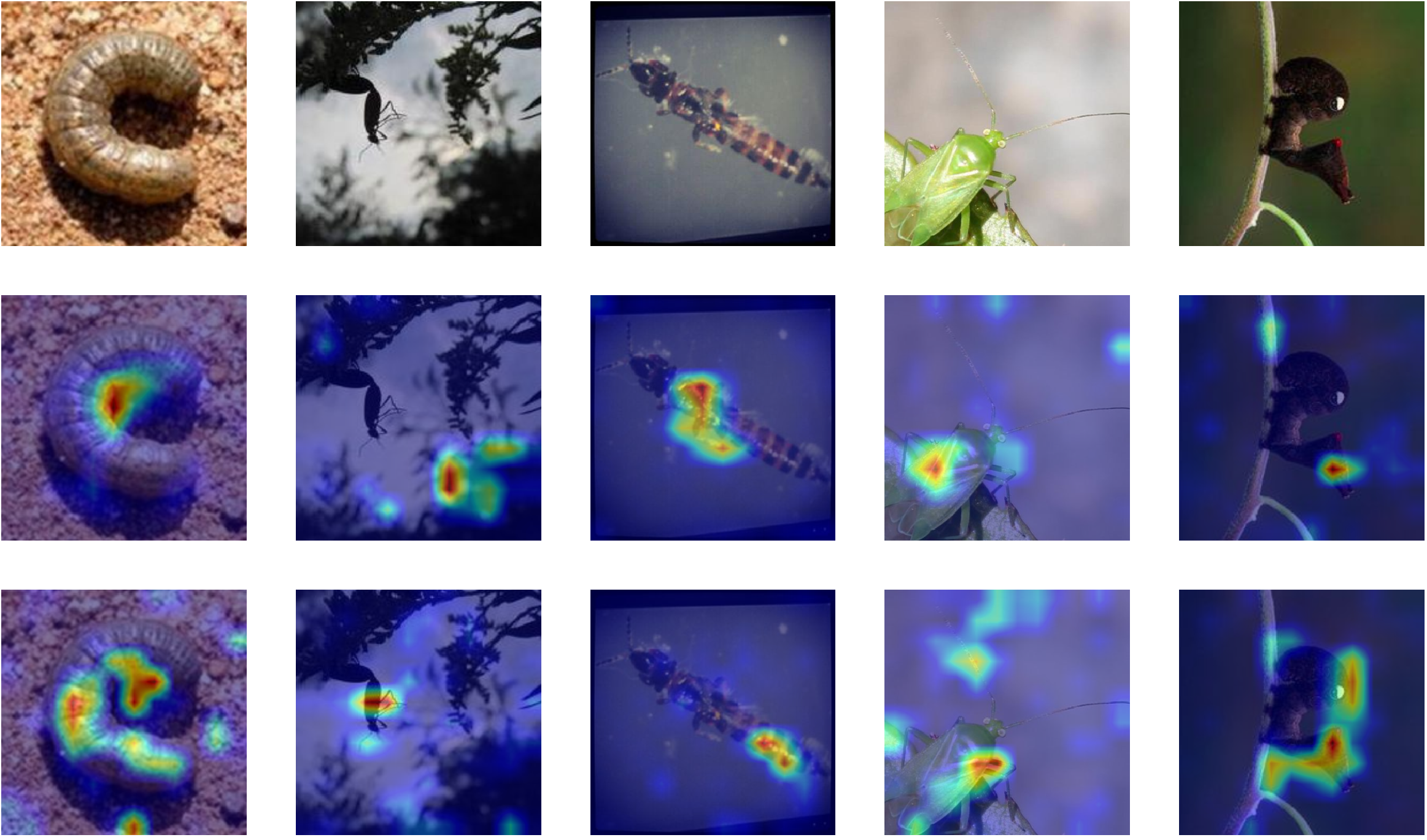}
	\caption{Visualization results of the ResNet-152 when applying and not applying DeWi. Each column represents the results of a specific insect. \textit{Top row:} input images. \textit{Middle row:} results of the baseline model. \textit{Bottom row:} results of the DeWi model.}
	\label{fig: grad_cam}
\end{figure}

We present additional visualizations to compare our DeWi with baseline residual networks. We select several cases where the baseline ResNet-152 misclassifies while our DeWi works correctly. We employ the Gradient-Weighted Class Activation Mapping (Grad-CAM) in~\cite{selvaraju2017grad} and provide class activation maps using the flowing gradients through the last core block. Visual results in Figure~\ref{fig: grad_cam} demonstrate that the DeWi model is better at focusing on meaningful features than the baseline ResNet-152.

\subsubsection{Feature space}
\begin{figure}[h]
\centering
\subfigure[ResNet152]{\includegraphics[height=60mm]{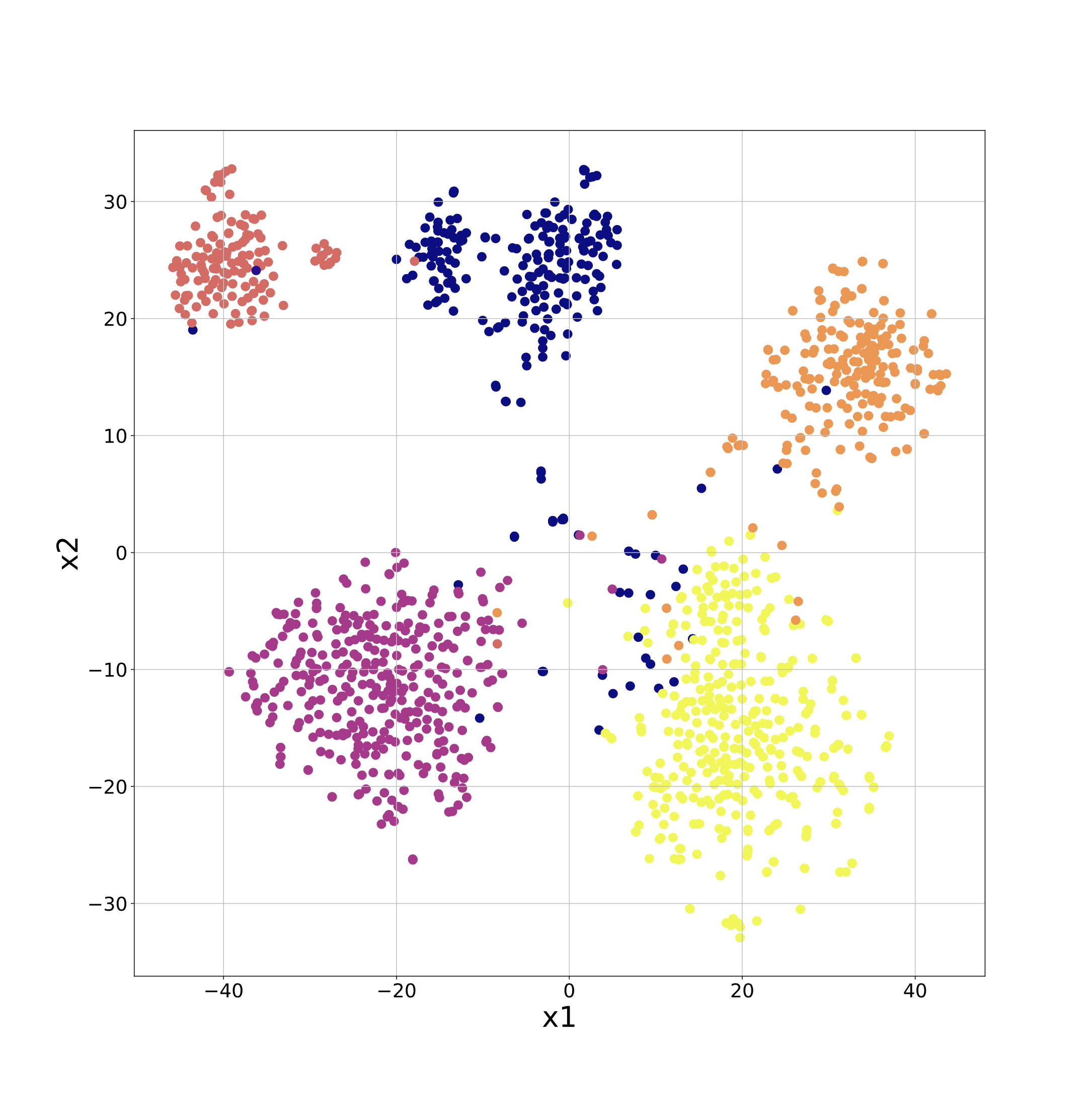}}
\subfigure[DeWi]{\includegraphics[height=60mm]{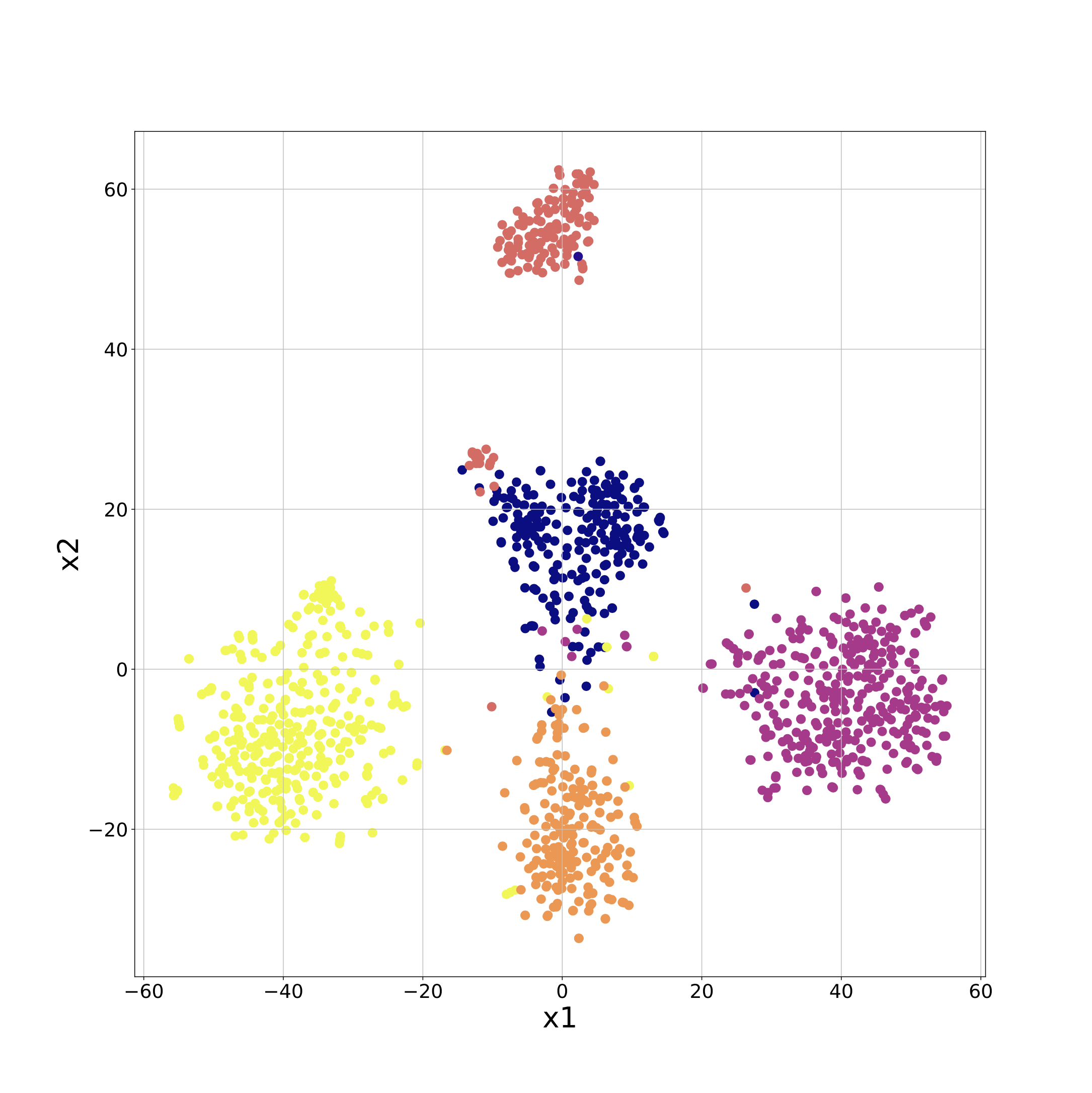}}
\caption{The t-SNE visualization results of the baseline ResNet-152 model and our DeWi model for five randomly selected classes.}
\label{fig: tsne}
\end{figure}

%\subsubsection{t-SNE visualization}
We further analyze the learning ability of our DeWi by utilizing the t-SNE algorithm~\cite{van2008visualizing} to visualize the feature space. In particular, we select test images of five random classes from the IP102 dataset and present the feature visualization of the baseline ResNet-152 and our DeWi on a 2D plane. For ResNet-152, we select the flattened output vector of the last core block. For our DeWi, we choose the concatenating feature vector of the multi-level extractor. The t-SNE is run for 1,000 iterations, with result is indicated in Figure~\ref{fig: tsne}. The visualization shows that DeWi is better at grouping features of the same classes and discriminating features of different classes, fortifying its effectiveness.

\subsection{Efficiency evaluation}
Previous works usually employ ensemble methods that integrate many different models~\cite{ung2021efficient,nanni2022high,ayan2020crop} or design heavy neural networks with many components that require many times of forwarding through the backbone~\cite{feng2022ms}. In contrast, DeWi is simple and exhibits a compact design. For that reason, our method is expected to be more efficient. To validate this point, we perform an evaluation to compare the running time of our method with the method proposed by Ung~\textit{et al.}~\cite{ung2021efficient}, which is the only one whose source code is publicly available. Specifically, we benchmark the average inference time (IT) of an image for each model on an NVIDIA GeForce RTX 3090Ti GPU. The results shown in Table~\ref{tab: efficiency} indicate that most of DeWi models are faster than~\cite{ung2021efficient} (with the only exception of Wide ResNet-101-2) while, as shown in Section~\ref{sec:baseline compare}, all DeWi models consistently outperform~\cite{ung2021efficient} by a large margin.

\begin{table}[h]
    \centering
    \resizebox{\textwidth}{!}{%
    \begin{tabular}{@{}cccccccccc@{}}
    \toprule
    & \makecell{Ung \textit{et al.}\\ \cite{ung2021efficient}} & \makecell{DeWi\\ ResNet-50} & \makecell{DeWi\\ ResNet-101} & \makecell{DeWi\\ ResNet-152} & \makecell{DeWi ResNext\\ -50 32x4D} & \makecell{DeWi ResNext\\ -101 32x8D} & \makecell{DeWi ResNext\\ -101 64x4D} & \makecell{DeWi Wide\\ ResNet-50-2} & \makecell{DeWi Wide\\ ResNet-101-2} \\ \midrule
    \makecell{IT (\SI{}{\milli\second})} & 76.23                               & \bf{60.93} & \underline{61.52} & 68.95 & 61.77 & 70.54 & 70.19 & 69.42 & 78.55               \\ \bottomrule
    \end{tabular}
    }
    \caption{Inference time comparison.}
    \label{tab: efficiency}
\end{table}

\subsection{Ablation Studies}\label{sec: ablation}
In this section, we conduct multiple ablation studies to investigate other aspects of DeWi. Firstly, we analyze the importance of each of DeWi's components. We then demonstrate the robustness of DeWi to the change in batch size. Next, we benchmark the performance of DeWi when replacing the triplet margin loss with other state-of-the-art loss functions in contrastive learning. Lastly, we compare DeWi with the self-supervised setting and discuss the outcomes. We use ResNet-152 as the underlying architecture for our ablation studies.

\textbf{The importance of each component.}
We conduct the first study on the importance of each DeWi's component and report the results in Table~\ref{tab: each component}. Note that for the networks that use triplet margin loss but without the multi-level feature extractor, the triplet margin loss is computed on the output of the last block of the residual backbone (a 2048-dim vector). For the model that applies Mixup augmentation but discards the triplet margin loss, we use Mixup for all mini-batch. The results indicate that triplet margin loss improves the performances of networks notably. Applying only the Mixup, on the other hand, does not improve the performance of the baseline model and the model using triplet loss without the multi-level extractor. In fact, their performances are even worse than the original ResNet-152. Mixup, however, improves the model's performance using the multi-level extractor and the triplet margin loss by 0.8 percentage points on the accuracy metric. The combination of all the components (DeWi model) achieves the best performance, illustrating the effectiveness of the overall framework.

\begin{table}[h]
\caption{The importance of each DeWi's component. ResNet-152 is chosen as the underlying network.~\emph{MP} denotes the models employing the multi-level feature extractor.}
\label{tab: each component}
\vskip 0.15in
\begin{tabular}{lccc}
\toprule
Method  &  Acc & mF1 & GM \\
\midrule
ResNet-152 & 75.15 & 68.06 & \underline{64.89} \\
ResNet-152 + Mixup & 75.07 & 66.68 & 61.19\\
ResNet-152 + Mixup + triplet loss & 75.11 & 67.55 & 63.58\\
ResNet-152 + MP & 74.86 & 67.77 & 64.70 \\
ResNet-152 + MP + triplet loss & \underline{75.64} & \underline{68.12} & 64.81 \\
ResNet-152 + Mixup + MP + triplet loss (DeWi) & \bf{76.44} & \bf{69.46} & \bf{65.07} \\
\bottomrule
\end{tabular}
\vskip -0.1in
\end{table}

\begin{figure}[h]
	\centering
	\includegraphics[width=0.7\linewidth]{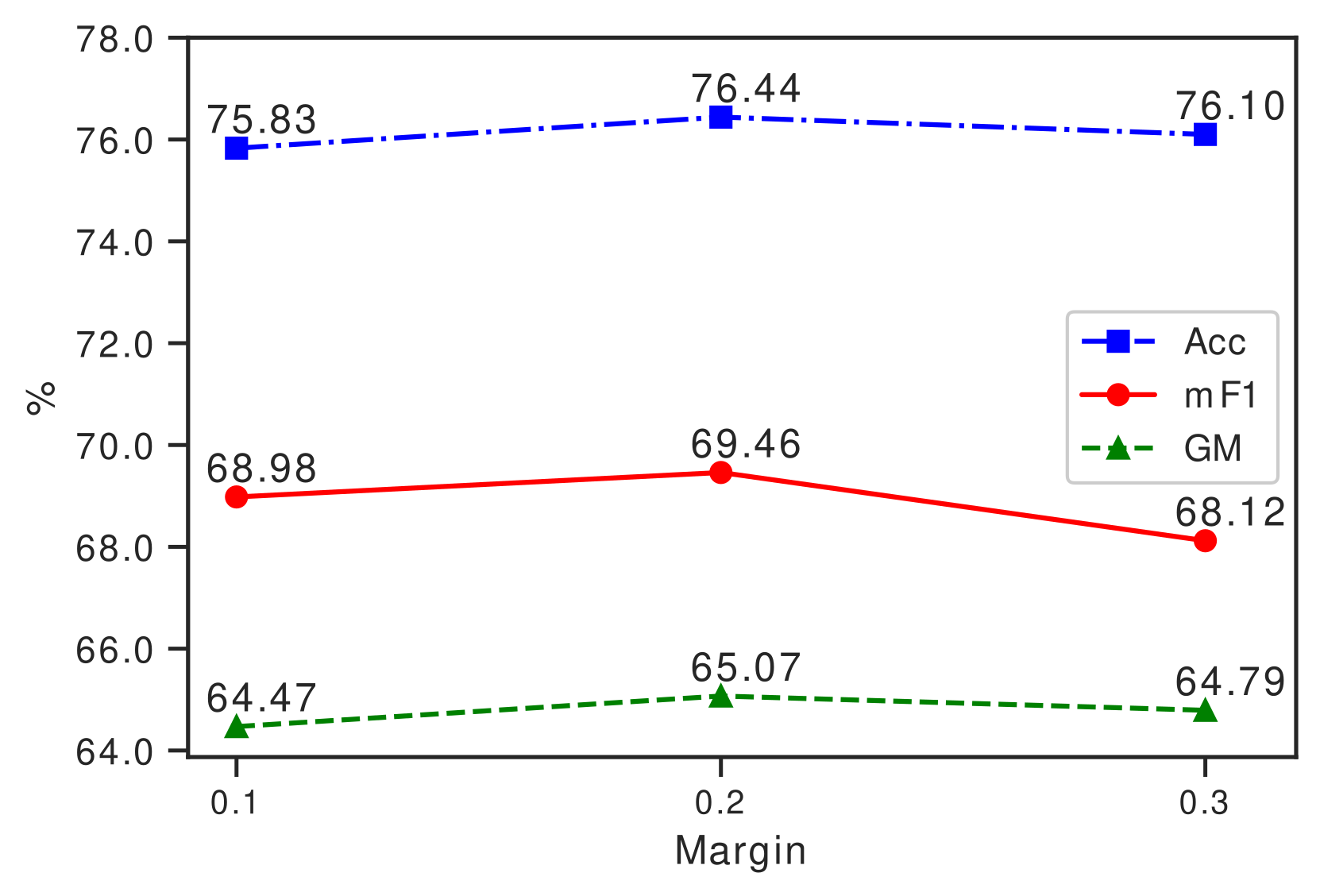}
	\caption{Performances of DeWi with different triplet loss margins.}
	\label{fig: margin}
\end{figure}

\textbf{Change in the margin hyper-parameter of triplet loss.}
The margin $m$ is an important hyper-parameter for the triplet margin loss~\cite{hermans2017defense}. For that reason, we report the performance of DeWi with different margin values of 0.1, 0.2, and 0.3. The results in Figure~\ref{fig: margin} show that the model with the margin of 0.2 achieves the best results on three metrics. We realize that when the margin is overly low, the triplet margin loss works ineffectively because of the low contrasting level. In contrast, when the margin is too high, it is challenging to optimize our loss. The appropriate value 0.2 of $m$ balances the learning of our model and leads the highest result.

\textbf{Different contrastive learning loss functions.}
We validate the effectiveness of the triplet margin loss function with the batch hard miner in our method by alternately replacing it with other two state-of-the-art contrastive learning loss functions, i.e., normalized temperature-scaled cross-entropy loss (NTXent loss)~\cite{chen2020simple} and Circle loss~\cite{sun2020circle}. For the NTXent loss $\mathcal{L}_{NTXent}$, we set the temperature hyperparameter to 0.07. The overall loss is computed by $\mathcal{L} = \mathcal{L}_{CE} + 0.1 \times \mathcal{L}_{NTXent}$. For the Circle loss $\mathcal{L}_{Circle}$, we set the relaxation factor to 0.4 and the scale factor to 80. The overall loss is then computed by $\mathcal{L} = \mathcal{L}_{CE} + 0.01 \times \mathcal{L}_{Circle}$. The results are shown in Table~\ref{tab: different loss}. Highest scores on all three metrics belong to the triplet margin loss. However, all three loss functions stably obtain good results, demonstrating the flexibility of our proposed framework.

\begin{table}[h]
\caption{Performances of DeWi with different contrastive learning loss functions.}
\label{tab: different loss}
\vskip 0.15in
\begin{tabular}{cccc}
\toprule
Batch size  &  Acc & mF1 & GM \\
\midrule
NTXent loss~\cite{chen2020simple} & 75.75 & 68.90 & 64.55 \\
Circle loss~\cite{sun2020circle} & \underline{76.18} & \underline{69.13} & \underline{64.88} \\
Triplet margin loss + batch hard miner~\cite{hermans2017defense} (ours) & \bf{76.44} & \bf{69.46} & \bf{65.07}\\
\bottomrule
\end{tabular}
\vskip -0.1in
\end{table}

\begin{figure}[h]
	\centering
	\includegraphics[width=0.7\linewidth]{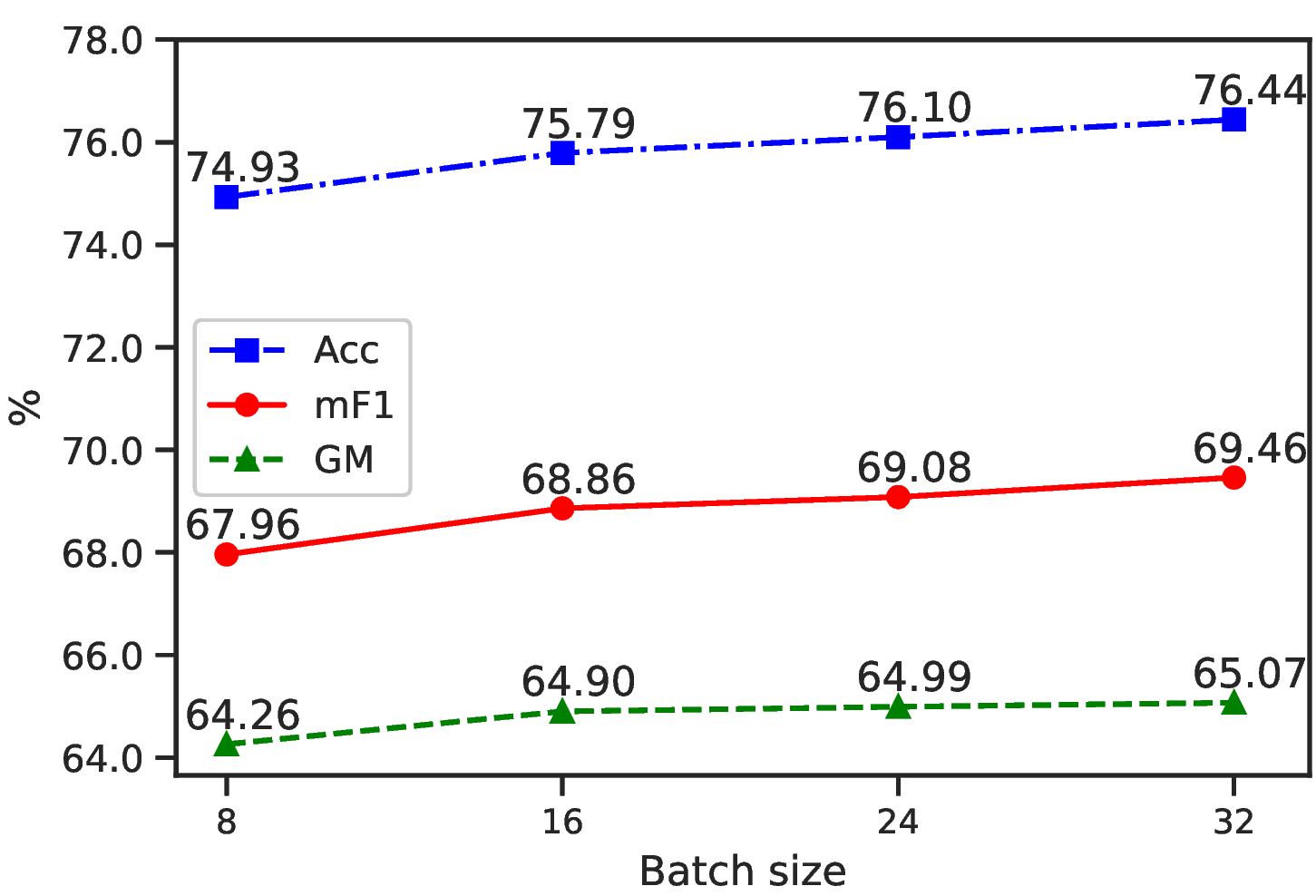}
	\caption{Performances of DeWi with different batch sizes.}
	\label{fig: batch size}
\end{figure}

\textbf{The robustness to batch size.}
The batch size hyper-parameter is an important factor for every contrastive learning method~\cite{jaiswal2020survey}. Therefore, we also study its effect on the performance of our proposed DeWi. In particular, we evaluate the performances of DeWi-assisted models with the batch size of ${32, 24, 16, 8}$. For the models with batch sizes of 24, 16, and 8, we set $1\mathrm{e}{-3}$ as the starting learning rate to speed up the learning process. Figure~\ref{fig: batch size} shows that all DeWi models trained by those different batch sizes obtain better results than other works (Table~\ref{tab: ip102 sota compare}).

\textbf{The effect of hyperparameter $\alpha$ in Mixup.}
In the Mixup process, we choose $\alpha=1.0$ so that the $\text{Beta}(\alpha, \alpha)$ distribution is equivalent to a uniform distribution. Our reason for this choice is that the uniform distribution is suitable for establishing different mixed combinations of input data with various ratios, consequently improving our model's generalization. To prove this point, we perform an additional ablation study on different values of $\alpha$. Particularly, we report the additional results on the IP102 dataset of our DeWi model with ResNet-152 backbone when $\alpha=0.5$ and $\alpha=2.0$, which lead to non-uniform distributions. Table~\ref{tab: alpha} shows that $\alpha=1.0$ yields the best performance, validating our choice.

\begin{table}[h]
\centering
\caption{Ablation study on $\alpha$.}
\label{tab: alpha}
\begin{tabular}{cccc}
\toprule
$\alpha$ & Acc & mF1 & GM \\ \midrule
0.5 & \underline{76.13} & \underline{69.25} & \underline{64.71} \\
1.0 & \textbf{76.44} & \textbf{69.46} & \textbf{65.07} \\
2.0 & 76.08 & 69.12 & 64.30 \\
\bottomrule
\end{tabular}
\end{table}

\textbf{Comparison with the pretext-based setting.}
We compare DeWi with a two-stage training setting where we train the model on a pre-text task in a self-supervised manner and then fine-tune the pre-trained network for the downstream classification task. In the self-supervised training stage, we remove the last linear layer of our DeWi and learn a pre-text task by optimizing the triplet margin loss on the concatenating output of the multi-level extractor. We train the model for 100 epochs with an initial learning rate of $3\mathrm{e}{-4}$. After pre-training, we re-insert the linear layer and train the downstream classification task for 50 more epochs. It is important to note that during the training on the main task, we freeze the pre-trained part and only update weights of the inserted linear layer.

Table~\ref{tab: self compare} shows the experimental results. First, it is shown that DeWi significantly outperforms the self-supervised pre-training method. Concretely, DeWi is largely better than its counterpart by 13.36, 15.74, and 20.26 percentage points on accuracy, mF1, and GM, respectively. We argue that the reason for the deficient performance of the self-supervised setting is that the batch size 32 is insufficiently small to learn the pre-text task. Like other self-supervised approaches, this setting is likely to require the batch size to be sufficiently large to achieve acceptable results. DeWi's one-stage training fashion, on the other hand, makes it work effectively even with small batch sizes. It also fosters the efficiency and applicability of DeWi in real-world applications.

\begin{table}[h]
\caption{Comparison between DeWi and the pretext-based setting.}
\label{tab: self compare}
\vskip 0.15in
\begin{tabular}{lccc}
\toprule
Method & Acc & mF1 & GM\\
\midrule
Self-supervised setting & 63.08 & 53.72 & 44.81\\
DeWi (ours) & \bf{76.44} & \bf{69.46} & \bf{65.07} \\
\bottomrule
\end{tabular}
\vskip -0.1in
\end{table}

\section{Conclusion}\label{sec: conclusion}
We have presented DeWi, novel learning assistance that can be applied to a wide range of CNN architectures. DeWi simultaneously facilitates the capabilities of learning discriminative features and generalizing to a wide range of insect pest classes. Our extensive experimental results showed that DeWi significantly improves the recognition capabilities of several CNN models, outperforming other state-of-the-art methods on insect pest recognition. We have also presented comprehensive ablation studies and experiments to validate the effectiveness and robustness of our method.

For future works, we plan to apply our DeWi for other CNN networks as well as other modern architectures like Vision Transformer~\cite{dosovitskiy2020image} to improve the performance. Besides, we aim to apply many other different contrastive learning loss functions and further propose our novel loss for the particular task of insect pest recognition. In addition, we will also investigate the performance of DeWi when adopting other image data augmentation methods, such as CutMix~\cite{yun2019cutmix}.

\bibliography{sn-bibliography}% common bib file
%% if required, the content of .bbl file can be included here once bbl is generated
%%\input sn-article.bbl

\end{document}